\documentclass[sigconf]{acmart}
\usepackage{graphicx}
\usepackage{multirow} % For \multirow
\usepackage{rotating} %
\usepackage[inkscapelatex=false]{svg}
\usepackage{enumitem}

\AtBeginDocument{%
  }

\copyrightyear{2024}
\acmYear{2024}
\setcopyright{rightsretained}
\acmConference[L@S '24]{Proceedings of the Eleventh ACM Conference on Learning @ Scale}{July 18--20, 2024}{Atlanta, GA, USA}
\acmBooktitle{Proceedings of the Eleventh ACM Conference on Learning @ Scale (L@S '24), July 18--20, 2024, Atlanta, GA, USA}
\acmDOI{10.1145/3657604.3664702}
\acmISBN{979-8-4007-0633-2/24/07}

\author{Momin N. Siddiqui}
\email{msiddiqui66@gatech.edu}
\orcid{0000-0003-1874-7789}
\affiliation{
  \institution{Georgia Institute of Technology}
  \streetaddress{North Avenue}
  \city{Atlanta}
  \state{GA}
  \country{USA}
  }

\author{Adit Gupta}
\email{ag3338@drexel.edu}
\orcid{0000-0002-1218-0630}
\affiliation{%
  \institution{Drexel University}
  \streetaddress{3230 Market Street}
  \city{Philadelphia} 
  \state{PA}
  \country{USA}
  }

\author{Jennifer M. Reddig}
\email{jreddig3@gatech.edu}
\orcid{0009-0000-6731-9401}
\affiliation{
  \institution{Georgia Institute of Technology}
  \streetaddress{North Avenue}
  \city{Atlanta}
  \state{GA}
  \country{USA}
  }

\author{Christopher J. MacLellan}
\email{cmaclellan3@gatech.edu}
\orcid{0000-0003-3084-5189}
\affiliation{
  \institution{Georgia Institute of Technology}
  \streetaddress{North Avenue}
  \city{Atlanta}
  \state{GA}
  \country{USA}
  }

\begin{document}

\title{ 
HTN-Based Tutors: A New Intelligent Tutoring Framework Based on Hierarchical Task Networks
}

\begin{abstract}
Intelligent tutors have shown success in delivering a personalized and adaptive learning experience. However, there exist challenges regarding the granularity of knowledge in existing frameworks and the resulting instructions they can provide. To address these issues, we propose HTN-based tutors, a new intelligent tutoring framework that represents expert models using Hierarchical Task Networks (HTNs). Like other tutoring frameworks, it allows flexible encoding of different problem-solving strategies while providing the additional benefit of a hierarchical knowledge organization. We leverage the latter to create tutors that can adapt the granularity of their scaffolding. This organization also aligns well with the compositional nature of skills. 

\end{abstract}


\maketitle

\section{Introduction}

Intelligent Tutoring Systems (ITSs) are computer programs that utilize AI techniques to provide personalized and adaptive learning \cite{nwana1990intelligent}. Several randomized controlled trials have demonstrated their effectiveness at improving student learning gains \cite{pane2014effectiveness, beal2007line}. Frameworks for intelligent tutoring include \textit{constraint-based} tutoring, which uses constraint-based modeling (CBM) to specify domain principles that every correct solution must follow \cite{ohlsson2016constraint}, and \textit{example-tracing} tutoring, which uses generalized traces of problem-solving behavior \cite{aleven2016example}. One of the more well-known frameworks is \textit{rule-based} tutoring, which uses production rules to represent the cognitive model, where every rule is tied to a student skill \cite{aleven2010rule}. Some examples of rule-based tutors adopted at scale are Cognitive Tutor/MATHia \cite{anderson1995cognitive, ritter2016mathia}, OATutor \cite{anastasopoulos2023introducing}, ASSISTments \cite{heffernan2014assistments} and Apprentice Tutor \cite{aialoe2024technologies}.

One line of research concerning intelligent tutors is the granularity of instructions. Granularity, in the context of intelligent tutors, refers to the amount of reasoning that is handled by the student internally on each step \cite{vanlehn2011relative}. Low granularity means that the system provides detailed, scaffolded steps to help guide the learner. In contrast, high granularity means that the system provides much less support---often just a single input box for the learner to enter the final answer. Research suggests that scaffolding should ideally be faded, shifting from low granularity to high granularity and reducing support as learners gain proficiency \cite{mcneill2006supporting, puntambekar2005tools}. However, limited research has explored ITS that dynamically changes the granularity of steps based on student skill level, functionality that we refer to as {\it adaptive scaffolding}. In most intelligent tutors, granularity is static. It is a feature of the user interface rather than encoded in the cognitive model, preventing adaption based on the learner's skill level. This gap necessitates formalizing a new expert model for intelligent tutors that can represent knowledge at different levels of granularity and better support adaptive scaffolding.

Across many existing systems, each skill the user is expected to acquire is represented independently and discretely \cite{vanlehn2001bayesian}. In the case of rule-based tutors, skills are represented as distinct production rules that activate when the rule's conditions are met. We argue that the representation of skills needs to better capture their compositional and hierarchical nature \cite{fischer1980theory}. 

To address the aforementioned challenges, we present a new framework using Hierarchical Task Networks (HTNs) that we refer to as {\it HTN-based} tutoring. HTNs are used in automated planning due to their human-like decomposition \cite{nau2003shop2}. We leverage this property of HTNs to encode different problem-solving strategies informed by the student's knowledge level. Each action is tied to meaningful abstractions representing knowledge components. HTN-based tutors capture a key characteristic of skills, they are built upon one another in compositional hierarchies. 

Our contribution is twofold. First, we propose a cognitive modeling framework that can deliver adaptive scaffolding, aligning with the learner's progress to reduce cognitive load for novices and challenge advanced learners. Second, by modeling the compositional nature of skill development, HTN-based tutors offer a more realistic and practical framework for learning complex subjects.

\section{Background}

In intelligent tutors, granularity is typically a characteristic of the user interface and not the cognitive model; granularity reflects the level of reasoning participants must engage in between interaction points \cite{vanlehn2011relative}. A larger grain size implies more reasoning with each interaction. Systems have been constructed with varying levels of granularity. Answer-based systems have the largest possible granularity, presenting learners with a problem and asking them to input the final answer without giving feedback to learners on intermediate steps. This requires the learner to do extensive mental reasoning for each response. In contrast, step-based and substep-based systems offer more frequent, detailed guidance. That is, they break problems down into multiple steps that the learner needs to perform, reducing the amount of independent reasoning required before each response. Step-based tutoring systems \cite{vanlehn2006behavior} aim to have users enter the steps they would naturally write down when solving a problem while showing their work. In contrast, substep-based tutors give scaffolding and feedback at a level of detail that is even finer than the steps students would typically exhibit when showing their work. Therefore, there is a missed opportunity for better delivery of substep tutoring using instructions, especially when informed by students' skill level.

\begin{figure*}
    \centering
    \includegraphics[width=0.75\textwidth]{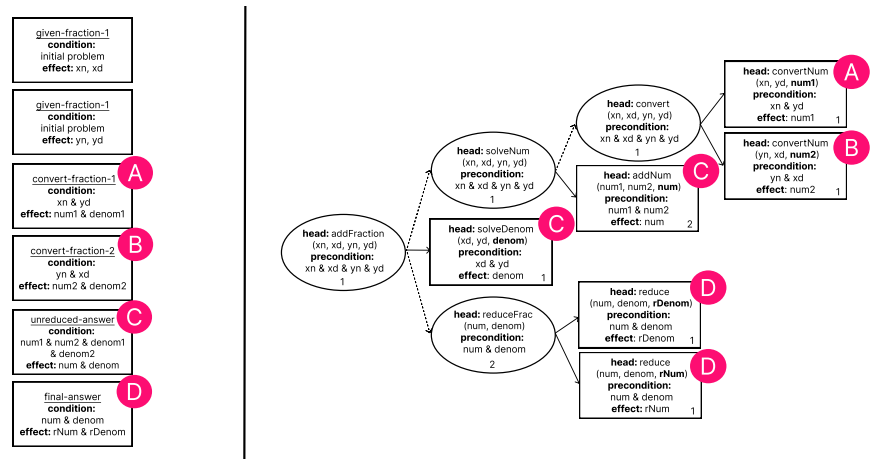} % adjust width of image
    \caption{A representation of fraction addition problem-solving knowledge in rule (left) and HTN formats (right), showing methods (ellipse) and operators (rectangle) for head tasks, with lettered callouts indicating equivalent steps in both frameworks.}
    \label{fig:one}
\end{figure*}

Hierarchical task networks (HTN) are popular for providing a robust framework for automated planning, an area of artificial intelligence research \cite{erol1994semantics}. HTNs are used in robotics, video games, and military simulations, among many other applications. Seminal systems such as SHOP and SHOP2 \cite{nau2003shop2} exemplify the effectiveness of HTNs in decomposing abstract tasks into concrete, manageable tasks through the use of Methods and Operators. Their hierarchical approach to task decomposition has a natural alignment with human cognitive skill organization, making it an intuitive framework for encoding task knowledge. This alignment facilitates a deeper understanding of planning as a cognitive function. While its potential as a cognitive modeling tool is evident \cite{hayes2016autonomously}, research on its use in intelligent tutors, and in the learning sciences more generally, is limited. Our proposed system builds upon the domain-independent framework of HTNs to create expert models for tutoring systems.

\section{HTN-Based Tutors}
In this section, we will outline the components of our proposed HTN-based tutoring framework. We will describe the state, the expert model, and how both support model tracing and tutoring.

\subsection{State}
The system utilizes a short-term working memory that contains an up-to-date description of the system's state. The working memory represents the state using facts, which are logical predicates that describe different elements in the tutor and how they relate. An example fact describing the value stored in the ``addFraction'' field might be:
\begin{verbatim}
    Fact(field='addFraction', value='1/2+1/4')
\end{verbatim}
Facts encode the details of the problem, its expected scaffolding, and the user’s expertise for each skill as calculated through knowledge tracing \cite{corbett1994knowledge}.
The knowledge of which step should be scaffolded can be dynamically inferred based on a user's skill levels using Axioms. Axioms are described in greater detail in the following section.

\subsection{Expert Model}
The expert model stores problem-solving strategies, including tasks, operators, methods, and axioms.
\subsubsection{\textbf{Tasks}}
A task represents an activity to perform.\footnote{Here we use the HTN definition of a {\it task} \cite{nau2003shop2}, which is different from typical tutoring system definition \cite{vanlehn2006behavior}.} Tasks can be performed using an Operator or a Method.  Each task has a definition associated with it, which we call the task head. The task head is essentially the name or identifier of the task. The task head corresponds to heads that appears in Operator or Method. In order to start problem-solving, a task has to be instantiated.

\subsubsection{\textbf{Operator}}
Operators represent primitive behaviors. Each operator has a head, preconditions, and effects. The preconditions are a partial description of a state that describes when the operator can be applied and The effects describe the actions to be take when applied. These usually result in some change to the state (e.g., updating a tutor field with a particular value). Operators are similar in many respects to the production rules in rule-based models. The key difference is that operators are only considered in the context of the target task that they are known to perform, as specified by their head, whereas all productions are considered in every state.

\subsubsection{\textbf{Method}}
Methods represent non-primitive behaviors. Like operators, each method has a head that defines the task it performs and preconditions that describe when it is applicable. However, unlike operators, methods have subtask decompositions instead of effects. There can exist multiple methods with the same head to describe different ways of decomposing a given task under different conditions. This allows the encoding of multiple strategies for performing a task. The method is essentially a higher-level abstraction that break down a task into simpler subtasks.

\subsubsection{\textbf{Axiom}}
Axioms are used to infer facts that are not explicitly asserted in the current state. Each axiom has a head and preconditions. However, unlike methods and operators, the head refers to a new fact that can be inferred when its preconditions are satisfied. 

\subsection{Model Tracing with HTNs}
Model tracing is a technique for inferring a student's mental operations given their observed behavior on a problem \cite{anderson1990cognitive}, enabling real-time and contextual feedback on the student's inferred state.

To support model tracing with HTN-based cognitive models, all problems start with a single higher-level task that is linked to the problem (e.g., reduce the logarithmic expression). The HTN model tracing system searches for a way to recursively decompose this task until it grounds out into operators that describe actions to be taken in the tutor. For each task, the system retrieves all the methods or operators that can achieve it and checks to see if their preconditions are satisfied. If any of the methods and/or operators are applicable, then it selects one (this choice is represented as an OR decomposition in Figure \ref{fig:two}). If a method is selected, then it decomposes the current task into a conjunction of subtasks, each of which the system tries to satisfy via recursive decomposition (the conjunction of subtasks to be satisfied is represented as an AND decomposition in Figure \ref{fig:two}). If an operator is selected, then it waits for the student to take action corresponding to the effects described by the operator. If the expected effects are not observed, then the system backtracks to find an alternative decomposition that matches the observed student action. When no decomposition matches, the system identifies the student action as incorrect. However, if the student takes an expected action, then model tracing continues and the system moves on to the next subtask that needs to be decomposed. Once all the tasks have been decomposed into observed student actions, then the system recognizes the problem as solved. 

In each state, the HTN-based system only considers methods that satisfy the current task/subtask. As a result, it evaluates fewer Methods/Operators than rule-based tutors, which evaluate all production rules in every state. Figure \ref{fig:one} compares a simplified rule-based expert model, comparable to what is used in a Cognitive Tutor \cite{anderson1995cognitive} (on the left) with a simplified HTN-based expert model (on the right). 

\section{Example}
\label{sec:example}
\begin{figure*}
    \centering
    \includegraphics[width=0.75\textwidth]{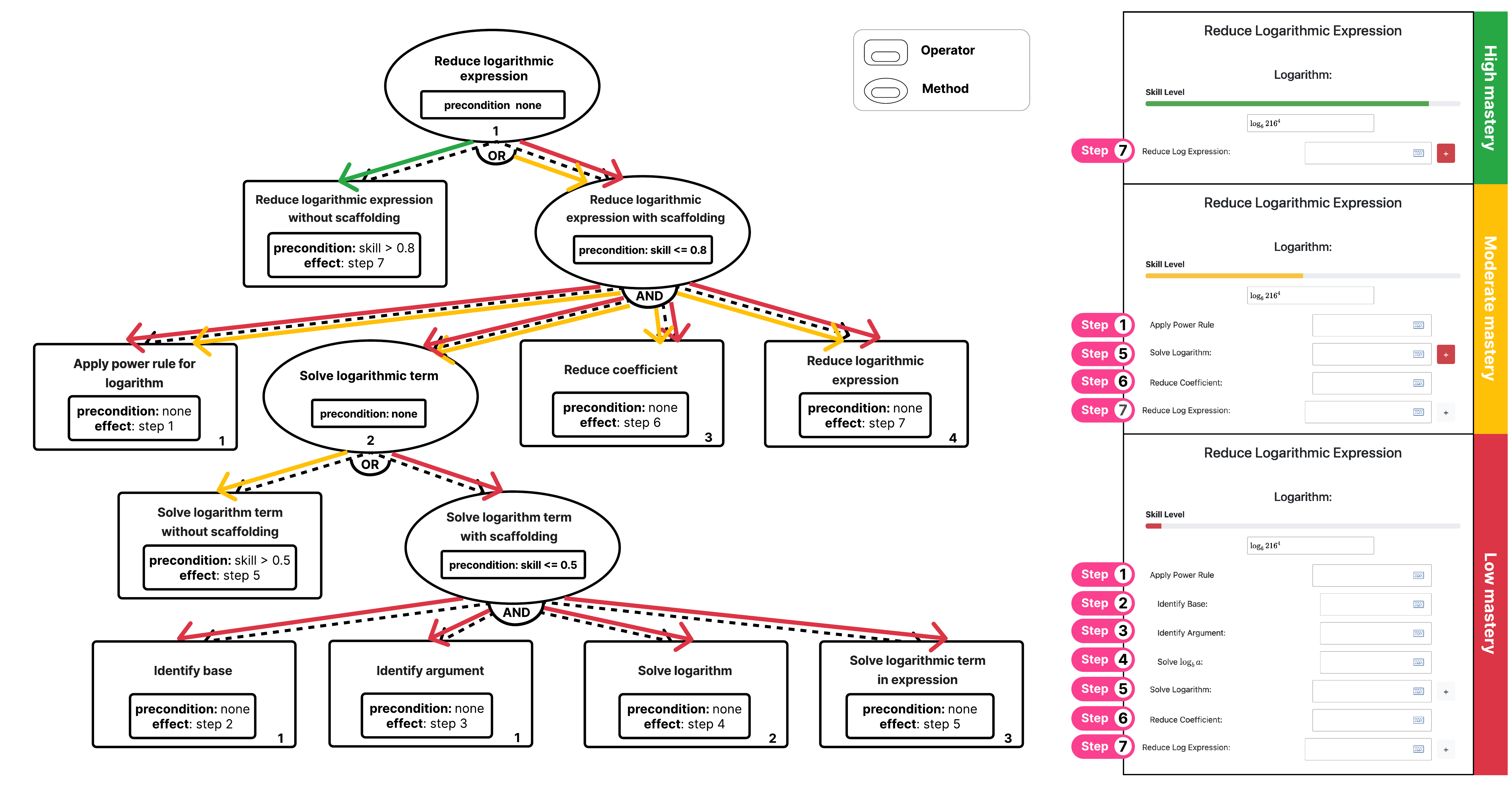} % adjust width of image
    \caption{An HTN-based tutor implementation for solving logarithmic expressions is shown. On the left, various paths for model tracing correspond to different skill levels: green for high-skill, yellow for moderate, and red for low-skill students.}
    \label{fig:two}
\end{figure*}

Having described the framework of HTN-based tutors, it is worthwhile to showcase an example in the context of a tutor. Figure \ref{fig:two} shows a tutor for reducing logarithmic expression. The tutor (right) displays different granularity levels based on the user's estimated knowledge of each component (e.g., as estimated via knowledge tracing \cite{corbett1994knowledge}). A student who is highly proficient (marked by the green progress bar) sees an interface similar to an answer-based tutor. A student with intermediate skills (yellow progress bar) sees an interface similar to step-based, while for a novice, the interface resembles a sub-step-based tutor, outlining even the steps that are more traditionally done mentally. 

Figure \ref{fig:two} also shows the expected model tracing path corresponding to each user (left). The numbering at the bottom of nodes represents the ordering sequence of their respective AND branches. Each path is marked with colors matching their progression. The green yields no scaffolding, while yellow and red have provide progressively more scaffolding in each case. The plus button on the right of the input field expands the scaffolding manually if the student would like more support. Red implies scaffolding is available, while grey implies scaffolding has already been expanded for that step.

\section{Discussion}

\subsection{Adaptive Scaffolding}
In intelligent tutors, the focus is typically on providing contextualized hints and personalized practice sequences rather than adaptive scaffolding \cite{anderson1995cognitive, anastasopoulos2023introducing}. These systems utilize two primary mechanisms: the outer loop and the inner loop \cite{vanlehn2006behavior}. The outer loop selects problems that learners have not yet mastered \cite{ritter2016mastery}, focusing their practice where it is needed rather than on skills they already know. The inner loop offers hints and immediate feedback directly through the tutor interface, tailoring to the specific context. Sometimes students with low skills do not have the metacognition to implement the hints and explanations they are given \cite{cusi2020re}. They require explicit problem-solving instruction and modeling of their problem-solving strategies to make progress. Our cognitive models enable alternative methods to administer adaptive scaffolding.

\subsubsection{\textbf{Granularity}}
Granularity is explicitly represented in HTN's task hierarchy, where higher-level tasks are less granular and lower-level tasks represent more granularity. This allows multiple subtask lists to be authored, catering to learners of varying skill levels, as seen in figure \ref{fig:two}. This approach not only addresses cognitive load but also enhances learning by dynamically adjusting the scaffolding based on the HTN model to meet the learner's evolving needs. Adaptive granularity enables the tutor to reduce the degrees of freedom \cite{wood1976role} for problem-solving while retaining higher-level task context by means of tailored instructions. Effective scaffolding should gradually fade as learners gain proficiency \cite{mcneill2006supporting}, allowing them to become more independent in their problem-solving. Our system progressively reduces the level of scaffolding provided. As learners advance, the tutor transitions to an answer-based format, fostering greater learner autonomy and reinforcing mastery. By giving students practice in a format that more closely resembles the format of problems on exams and tests, we hypothesize our tutors will yield better performance gains under testing conditions than typically seen with ITS \cite{kulik2016effectiveness}.

\subsubsection{\textbf{Strategy Recognition}}
It is valuable for students to receive feedback on the strategies they use, as this enables them to meta-cognitively assess their problem-solving. Due to strategy being encoded directly in HTN methods, our model tracing system can identify the approach taken by students and provide feedback on their strategic choices. This should help improve student's current understanding and enhance their outcomes \cite{wood1976role}.

\subsection{Representation Power}
Skills are compositional in nature \cite{fischer1980theory}, serving as building blocks for higher-level skills. In contrast to production rule engines, which process rules individually without a hierarchical knowledge structure, our HTN-based approach organizes knowledge hierarchically, reflecting the natural organization of skills. The flexible nature of the HTN-based tutor allows for the reusability of methods and operators while not compromising the hierarchy of skills. Moreover, the tasks in the HTN do not have to be strictly sequential and can have interleaved dependency. This makes it easier to encode complex problems where tasks do not follow a linear sequence of steps (e.g., where the subtasks can be performed in any order).

\section{Limitation \& Future Work}

While HTN-based tutors show promise, they have yet to be tested and integrated within a large-scale deployment. Integrating our novel cognitive modeling approach within a platform to support standard inner and outer loop features would allow for a more holistic evaluation. We plan to investigate the following  questions:
\begin{enumerate}
    \item Does adaptive scaffolding, through adjusting granularity, improve student learning?
    \item What is the optimal strategy for adaptive granularity?
\end{enumerate}
To address the first question, we will compare the performance of intelligent tutors with static granularity to HTN-based tutors with adaptive granularity through controlled experiments. Students will be randomly assigned to one of three groups: an HTN-based tutor with static scaffolding, one with adaptive scaffolding, or a control group without tutoring. We will evaluate effectiveness using learning gains measured by pre- and post-tests.
For the second question, we will use a similar design with two treatment groups: one where granularity follows a U-shaped curve (starting large, decreasing, then increasing) and another where granularity follows a sigmoid curve (starting small and increasing). Outcomes will be assessed using pre- and post-tests.

\section{Conclusion}
To address the limitations of existing tutoring systems, we present an alternative framework for tutor cognitive models that uses HTNs. HTN-based tutors provide better personalization by enabling adaptive scaffolding functionalities that have thus far been under explored. The hierarchical nature of the system enables reusability of skills across tutors while maintaining their hierarchical association. It allows the composition of skills to formulate new higher-level skills, enabling better knowledge compilation \cite{fischer1980theory}. The human-like encoding of skills opens the opportunity to incorporate several features that will help us deliver a better experience for learners.

\section{Acknowledgement}
This project is supported by National Science Foundation under Grant No. 2247790 and Grant No. 2112532. Any opinions, findings, and conclusions or recommendations expressed in this material are those of the author(s) and do not necessarily reflect the views of the National Science Foundation.

\bibliographystyle{ACM-Reference-Format}
\bibliography{main}

\end{document}